\documentclass{article}
\usepackage[final,nonatbib]{nips_2016}
\usepackage[utf8]{inputenc} 
\usepackage[T1]{fontenc}    
\usepackage{graphicx}
\usepackage[numbers]{natbib}
\usepackage{cleveref}
\usepackage{url}

\title{ResearchDoom and CocoDoom:\\Learning Computer Vision with Games}
\author{%
A. Mahendran \\
Department of Engineering Science\\
University of Oxford \\
Oxford UK \\
\texttt{aravindh@robots.ox.ac.uk} 
\And H. Bilen \\
Department of Engineering Science\\
University of Oxford \\
Oxford UK \\
\texttt{hbilen@robots.ox.ac.uk} 
\And J. F. Henriques \\
Department of Engineering Science\\
University of Oxford \\
Oxford UK \\
\texttt{joao@robots.ox.ac.uk} 
\And A. Vedaldi \\
Department of Engineering Science\\
University of Oxford \\
Oxford UK \\
\texttt{vedaldi@robots.ox.ac.uk}}

\begin{document}
\maketitle
\begin{abstract}
In this short note we introduce ResearchDoom, an implementation of the Doom first-person shooter that can extract detailed metadata from the game. We also introduce the CocoDoom dataset, a collection of pre-recorded data extracted from Doom gaming sessions along with annotations in the MS Coco format. ResearchDoom and CocoDoom can be used to train and evaluate a variety of computer vision methods such as object recognition, detection and segmentation at the level of instances and categories, tracking, ego-motion estimation, monocular depth estimation and scene segmentation. The code and data are available at \url{http://www.robots.ox.ac.uk/~vgg/research/researchdoom}.
\end{abstract}

\section{Introduction}\label{s:intro}

As data-hungry methods such as deep neural networks become predominant in most  areas of computer vision, the community faces the challenge of obtaining sufficiently large quantities of supervised data to train models. One increasingly popular and effective approach is to use synthetic data~\cite{goodfellow14multi-digit,jaderberg14synthetic,gupta16synthetic,hong16faces}. While this data can be created ad-hoc, research in reinforcement learning has recently built on the idea of using video games as off-the-shelf virtual environments suitable to learn agents~\cite{oh15action-conditional,mnih13playing,brockman16openai}. This approach is rather powerful as, while video games may not be entirely realistic, they capture many of the fundamental nuances and challenges of real-life perception. This is especially true for modern 3D games. We believe that these ideas can be applied well beyond reinforcement learning, and that video games can be used to train and evaluate in a direct manner a wide variety of computer vision algorithms.

In order to explore this idea, in this paper we propose to extract richly-annotated data from Doom, one of the early examples of 3D games. In order to do so, we introduce a modification of the open-source Doom engine. This variant, which we call \emph{ResearchDoom} (\cref{s:rdm-engine}), is able to extract a wealth of geometric and semantic metadata during gaming sessions. Second, in order to lower as much as possible the barrier of entry for other computer vision researchers, we also provide pre-computed Doom data with full annotations, as well as a subset of the annotations using the Microsoft Coco format~\cite{lin14microsoft}, including well defined training, validation, and test split to simplify comparing results. We call this dataset \emph{CocoDoom} (\cref{s:rdm-coco}).

We open source both ResearchDoom and CocoDoom and make them available at \url{http://www.robos.ox.ac.uk/~vgg/research/researchdoom} for other researchers to use.

\subsection{Related work}\label{s:related}

As noted above, several researchers have been proposing to use computer graphics and simulated data for learning in computer vision and reinforcement learning. The work that is probably most related to ours is \emph{VizDoom}, a modified version of the Doom engine that allows interfacing the game to machine learning algorithms for the purpose of reinforcement learning. ResearchDoom, which was developed partially concurrently to VizDoom, rather than providing an interface to control the game as needed in reinforcement learning, it focuses on extracting abundant metadata from recorded gaming sessions, including extracting depth information, object instance and class segmentations, as well as other data such as egomotion. The idea is that such annotated data can be used \emph{directly} in a large number of computer vision tasks even without reinforcement learning. Example problems that can be targeted by using this data include object recognition, detection, and tracking, instance and semantic segmentation, monocular depth estimation, and ego-motion estimation. While ResearchDoom was developed independently of VizDoom, in the future we hope to be able to port ResearchDoom functionalities to VizDoom.

Another recent effort similar to ResearchDoom is Unreal CV~\cite{qiu16unrealcv:}. This engine can also extract metadata similar to ResearchDoom and, by building on the Unreal engine, can potentially be applicable to a huge variety of modern video games. ResearchDoom is much simpler than Unreal CV as it applies to a single and relatively old game. While Doom incorporates several limitations, such as the fact that the camera can only rotate around the vertical axis, the data is nevertheless fairly complex. Furthermore, Doom provides a relatively restricted and consistent world form which data can be extracted for experiments. We leverage the latter fact to define and provide abundant pre-computed data with detailed metadata as well as with annotations using the Microsoft Coco format.

Another related project is the OpenAI Gym~\cite{brockman16openai}, a collection of simulations and virtual environments, including video games such as Doom, for research in reinforcement learning by OpenAI.

\begin{figure}
\includegraphics[width=\textwidth]{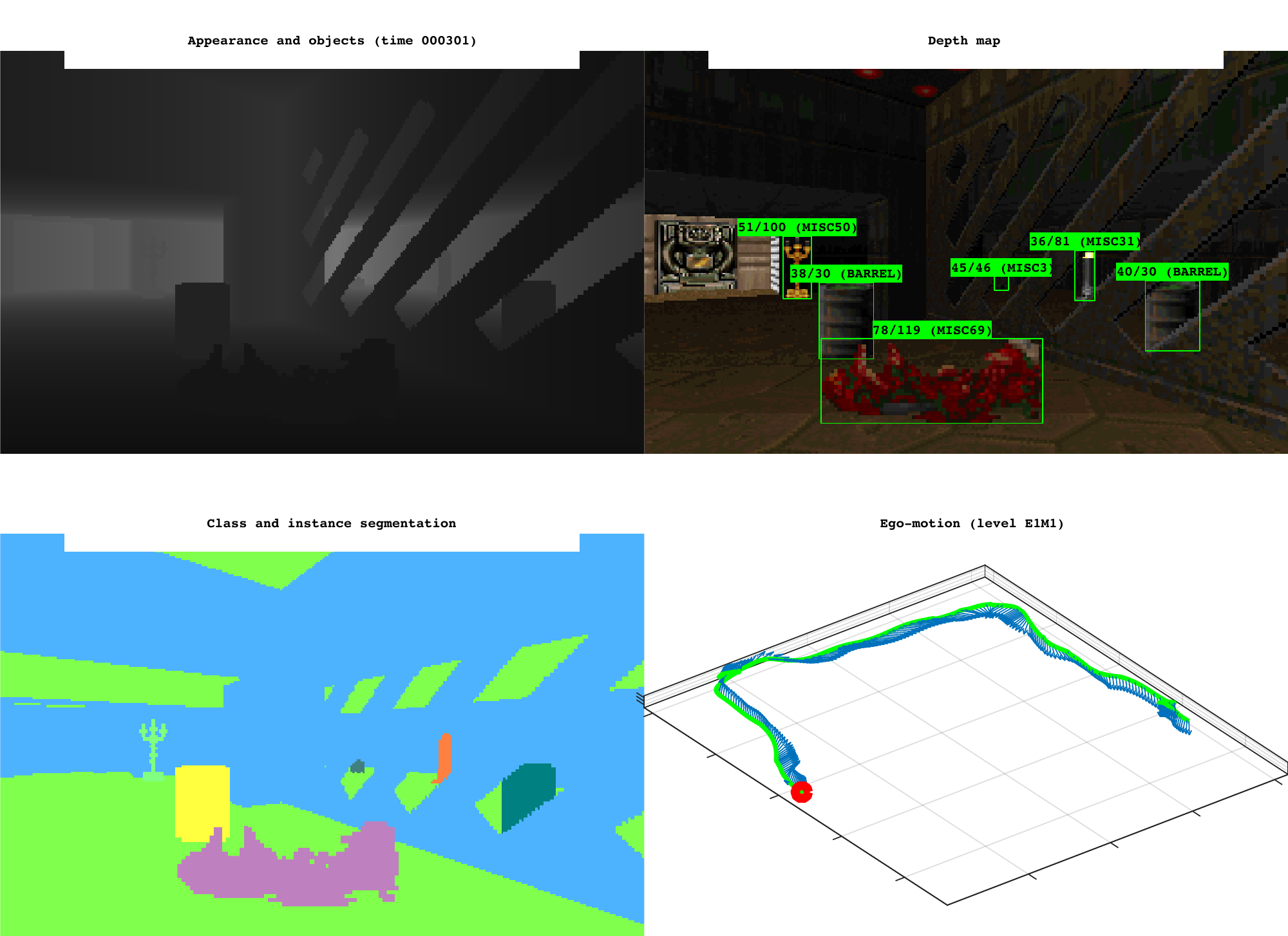}
\caption{\textbf{ResearchDoom data and metadata.} Top-left: depth map. Top-right: appearance with bounding box annotations for objects (each object is labelled with an instance identifier and a class identifier). Bottom-left: object masks as well as vertical/horizontal surfaces segmentation. Bottom-right: ego-motion.}\label{f:rdm}
\end{figure}

\section{The ResearchDoom engine}\label{s:rdm-engine}

The \emph{ResearchDoom engine} is a minimal modification of the Chocolate Doom\footnote{\url{https://www.chocolate-doom.org/}} engine. It allows extracting various types of data and metadata from the game, including appearance, egomotion, depth, and object masks. By using ResearchDoom it is possible to cheaply generate large quantities of images from this virtual 3D environment annotated in great detail and without errors. 

The ResearchDoom engine is typically used to extract data from pre-recorded playthroughs. It is deterministic, in the sense that, given these game files, all the data can be reproduced exactly.

Time in Doom is measured in terms of the so-called \emph{tics}. Each tic produces exactly a frame, and for each frame ResearchDoom saves three PNG files, containing the appearance information, the depth map, and the object masks. The engine also writes a log of game events in a text file which can be used, among other things, to recover the location and orientation of the player in the 3D world as well as the category of each object instance. The detailed format of these files is described in the ResearchDoom homepage.

\section{The CocoDoom data}\label{s:rdm-coco}

\begin{figure}[t]
	\includegraphics{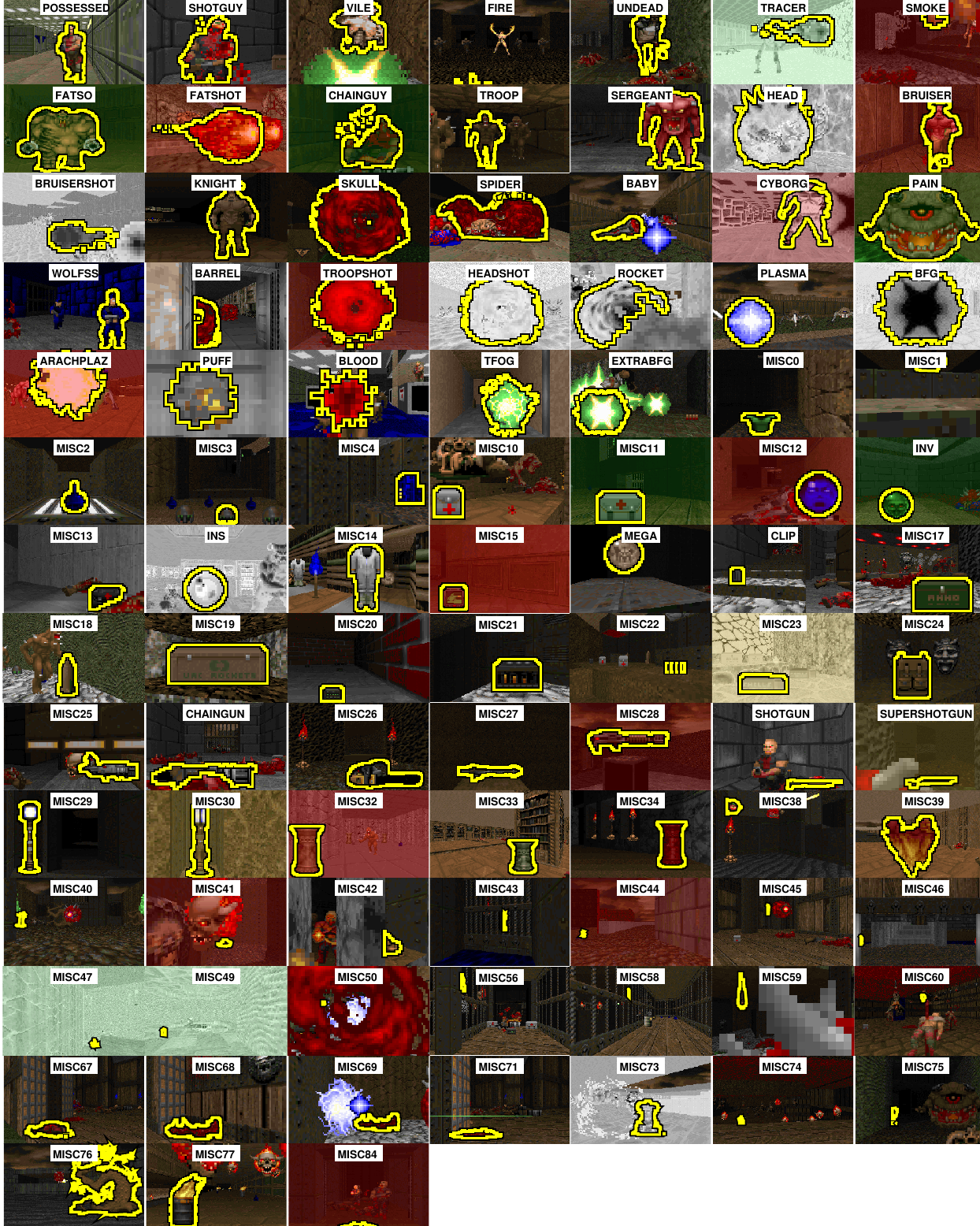}
	\caption{\textbf{The 94 CocoDoom categories.}}\label{f:cats}
\end{figure}

As a simple point of entry for computer vision researchers, we also provide extensive pre-recorded data. The data contains the annotations produced by the ResearchDoom engine, as described in~\cref{s:rdm-engine}. It also contains object annotations in the same format used by \emph{Microsoft Coco}~\cite{lin14microsoft}. Therefore, it is a plug-and-play resource for computer vision systems that use this format.

The data itself was generated from three complete playthroughs of the Doom II version of the game. The game itself consists of 32 maps or levels, divided in four episodes. We consider two different splits of the data in training, validation, and test sets as follows. The \emph{run} split uses the first playthrough for training, the second for validation, and the last for testing. The \emph{episode} split uses the first two episodes for training, the third for validation, and the fourth for testing. The \emph{episode} split is therefore harder as different game maps are contained in the different splits.

All the frames were extracted from each run. Due to the large quantity of data produced in this manner, we also consider a \emph{standard} and a \emph{full} subset. The standard subset contains only appearance images and Microsoft Coco annotations for the objects (which means that their boundaries are approximated by polygons) whereas the full data contains depth information and pixel-level object segmentations. In the MS Coco annotations, objects with an area smaller than 30 pixels are discarded. The standard subset also contains only 1/5 of all the possible frames. Categories with less than 100 representative images in the standard training data split where removed, remaining with 94 object types (\cref{f:cats}). Overall, the standard subset contains roughly 80K images and 300K object instance annotations, whereas the full data contains roughly 500K images and 1.4M object instance annotations.
 
\section{Conclusions}

In this short note we have introduced ResearchDoom and CocoDoom, respectively software to extract annotated data from the Doom game engine and a pre-computed collection of annotated gaming data suitable for training and evaluating a variety of computer vision algorithms. This effort is motivated by our belief that video game data, as well as computer graphics, synthetic, and augmented data in general, can provide a wealth of information for training computer vision systems. While the utility of simulated environments is obvious for reinforcement learning applications, until the problem of unsupervised learning is finally solved, such data can also be extremely useful in direct computer vision problems due to the ability of extracting automatically a very detailed supervisory signal.

In the future, we would like to merge ResearchDoom with the VizDoom codebase due to the technical merits and advantages of the latter. At the same time, we think that providing pre-computed gaming data and corresponding benchmarks in familiar formats can encourage the adoption of gaming data by a much larger portion of the computer vision community. This is the idea behind CocoDoom and for its future evolutions we hope to be able to leverage the power of more open-ended gaming engines such as Unreal CV.

\bibliography{references}

\begin{thebibliography}{1}

\bibitem{brockman16openai}
Greg Brockman, Vicki Cheung, Ludwig Pettersson, Jonas Schneider, John Schulman,
  Jie Tang, and Wojciech Zaremba.
\newblock Openai gym.
\newblock {\em CoRR}, abs/1606.01540, 2016.

\bibitem{goodfellow14multi-digit}
Ian~J Goodfellow, Yaroslav Bulatov, Julian Ibarz, Sacha Arnoud, and Vinay Shet.
\newblock Multi-digit number recognition from street view imagery using deep
  convolutional neural networks.
\newblock In {\em Proc. {ICLR}}, 2014.

\bibitem{gupta16synthetic}
Ankush Gupta, Andrea Vedaldi, and Andrew Zisserman.
\newblock Synthetic data for text localisation in natural images.
\newblock {\em Proc. {CVPR}}, 2016.

\bibitem{hong16faces}
Y.~Hong, R.~Arandjelovi\'c, and A.~Zisserman.
\newblock Faces in places: Compound query retrieval.
\newblock In {\em Proc. {BMVC}}, 2016.

\bibitem{jaderberg14synthetic}
M.~Jaderberg, K.~Simonyan, A.~Vedaldi, and A.~Zisserman.
\newblock Synthetic data and artificial neural networks for natural scene text
  recognition.
\newblock In {\em {NIPS} Deep Learning Workshop}, 2014.

\bibitem{lin14microsoft}
Tsung{-}Yi Lin, Michael Maire, Serge Belongie, James Hays, Pietro Perona, Deva
  Ramanan, Piotr Doll{\'{a}}r, and C.~Lawrence Zitnick.
\newblock Microsoft {COCO:} common objects in context.
\newblock {\em CoRR}, abs/1405.0312, 2014.

\bibitem{mnih13playing}
Volodymyr Mnih, Koray Kavukcuoglu, David Silver, Alex Graves, Ioannis
  Antonoglou, Daan Wierstra, and Martin~A. Riedmiller.
\newblock Playing atari with deep reinforcement learning.
\newblock {\em CoRR}, abs/1312.5602, 2013.

\bibitem{oh15action-conditional}
Junhyuk Oh, Xiaoxiao Guo, Honglak Lee, Richard~L Lewis, and Satinder Singh.
\newblock Action-conditional video prediction using deep networks in atari
  games.
\newblock In {\em Advances in Neural Information Processing Systems}, pages
  2863--2871, 2015.

\bibitem{qiu16unrealcv:}
Weichao Qiu and Alan Yuille.
\newblock Unrealcv: Connecting computer vision to unreal engine.
\newblock {\em arXiv preprint arXiv:1609.01326}, 2016.

\end{thebibliography}
\end{document}